# Bridging Artificial Intelligence and Data Assimilation: The Data-driven Ensemble Forecasting System ClimaX-LETKF

Akira Takeshima[1], Kenta Shiraishi[2], Atsushi Okazaki[1,3], Tadashi Tsuyuki[1,4], and Shunji Kotsuki[1,3,5]

[1] *Center for Environmental Remote Sensing, Chiba University, Chiba, Japan*

[2] *Graduate School of Science and Engineering, Chiba University, Chiba, Japan*

[3] *Institute for Advanced Academic Research, Chiba University, Chiba, Japan*

[4] *Observation and Data Assimilation Research Department, Meteorological Research Institute*

[5] *Research Institute of Disaster Medicine, Chiba University, Chiba, Japan*

Corresponding authors:

Akira Takeshima, Center for Environmental Remote Sensing, Chiba University, Chiba 263-8522, Japan. E-mail: takeshima@chiba-u.jp.

Shunji Kotsuki, Institute for Advanced Academic Research, Chiba University, Chiba 263-8522, Japan. E-mail: shunji.kotsuki@chiba-u.jp

## Abstract

While machine learning-based weather prediction (MLWP) has achieved significant advancements, research on assimilating real observations or ensemble forecasts within MLWP models remains limited. We introduce ClimaX-LETKF, the first purely data-driven ML-based ensemble weather forecasting system. It operates stably over multiple years, independently of numerical weather prediction (NWP) models, by assimilating the NCEP ADP Global Upper Air and Surface Weather Observations. The system demonstrates greater stability and accuracy with relaxation to prior perturbation (RTPP) than with relaxation to prior spread (RTPS), while NWP models tend to be more stable with RTPS. RTPP replaces an analysis perturbation with a weighted blend of analysis and background perturbations, whereas RTPS simply rescales the analysis perturbation. Our experiments reveal that MLWP models are less capable of restoring the atmospheric field to its attractor than NWP models. This work provides valuable insights for enhancing MLWP ensemble forecasting systems and represents a substantial step toward their practical applications.

## 1. Introduction

Machine learning (ML) has rapidly advanced the field of weather prediction, with ML-based weather prediction (MLWP) models now demonstrating performance comparable to, and in some instances exceeding, established operational forecasting systems such as the Integrated Forecasting System (IFS; Wedi et al. 2015) and the Global Forecast System (GFS; NCEP 2021) in medium-range forecasts (Bi et al. 2022; Bodnar et al. 2024; Chen, Kang et al. 2023; Chen, L. et al. 2023; Keisler et al. 2022; Lam et al. 2023; Price et al. 2023). Recent breakthroughs include the development of two end-to-end data-driven systems: FengWu-Adas (Chen, Kun et al. 2023) and Aardvark Weather (Allen et al. 2025). FengWu-Adas produces forecasts by integrating the MLWP model FengWu (Chen, Kang et al. 2023) with the data assimilation (DA) model Adas (Chen, Kun et al. 2023). Aardvark Weather, in contrast, directly estimates the initial atmospheric latent state from observation data, from which it generates subsequent predictions.

Most MLWP models developed to date have primarily been trained using global reanalysis data such as ERA5 (Hersbach et al. 2020). While they can infer future atmospheric states from given initial conditions (i.e., analysis fields), they cannot produce their initial conditions. For example, the European Centre for Medium-Range Weather Forecasts (ECMWF) implemented the AI Forecasting System operationally in February 2025, in which initial conditions are provided by a conventional DA system using the IFS numerical weather prediction (NWP) model (Lang et al. 2024). Even when observation data are assimilated into MLWP models, it is generally difficult for MLWP models to operate stably, especially when using real observations.

To achieve stable DA cycles, it is essential to explore more suitable MLWP models for data assimilation and to investigate appropriate DA techniques tailored to them. While some studies have examined the differences between MLWP and NWP models and developed MLWP models incorporating physical constraints, research specifically focused on DA remains very limited. The integration of MLWP and DA has been studied using simple models such as the 40-variable Lorenz 96 model (Lorenz 1996), prediction model of very limited variables in the atmosphere, or synthetic observations rather than real ones for DA (e.g., Adrian et al. 2024; Chattopadhyay et al. 2022; Hamilton et al. 2016; Kotsuki et al. 2025; Li et al. 2024; Penny et al. 2022). While Slivinski et al. (2025) recently succeeded in stabilizing DA cycle with the pure MLWP model with suppression of high-frequency noise with a Gaussian spectral filter, further studies are needed to enable stable data-driven weather predictions with assimilation of real observation data.

It is well known that the weather prediction accuracy is strongly influenced not only by data assimilation method, such as EnKF or 4D-Var, but also by the choice and tuning of covariance inflation methods. In this study, we investigated how the choice and tuning of covariance inflation method affect the stability of the DA cycle when using MLWP models. To this end, we developed ClimaX-LETKF, a purely data-driven ML-based ensemble weather forecasting system. Here, ClimaX-LETKF generates initial states by assimilating real observations (Figure 1), independently of NWP models. The system's performance exhibited high sensitivity to the settings of covariance inflation, underscoring their critical role in maintaining stable DA cycle in MLWP models. Our results represent a significant advancement toward the development of future weather prediction systems with MLWP models.

The remainder of this paper is organized as follows. Section 2 provides the ClimaX-LETKF system, Section 3 provides results of DA experiments, Section 4 provides a discussion and some concluding remarks.

## 2. ClimaX-LETKF system

### 2.1. MLWP model ClimaX

A foundation MLWP model, ClimaX (Nguyen et al. 2023), is used. This model is adaptable to a variety of weather- and climate-related tasks with appropriate training. We trained it from scratch on WeatherBench (Rasp et al. 2020), a standardized subset derived from ERA5, using 6-hourly data from 00:00 UTC on 1 January, 2006, to 18:00 UTC on 31 December, 2015. WeatherBench includes selected variables (e.g., geopotential at 500 hPa, temperature at 850 hPa, etc.) regridded to a lower resolution (e.g., 5.625°) and formatted to facilitate comparison across different data-driven forecasting models. Our system aimed to generate predictions 6 h ahead of the initial state and is optimized to minimize the MSE between predictions and the reference values for the following nine output variables: zonal wind, meridional wind, temperature, specific humidity, geopotential for seven atmospheric layers (925, 850, 700, 600, 500, 250 and 50 hPa, respectively), surface pressure, zonal wind and meridional wind at 10 m height, and temperature at 2 m height. The spatial resolution is 5.625°×5.625° (64 × 32 grid points) in the longitude and latitude directions, corresponding to the low-resolution version of WeatherBench. See Kotsuki et al. (2025) and Supplement 3 for more details on how the model was trained.

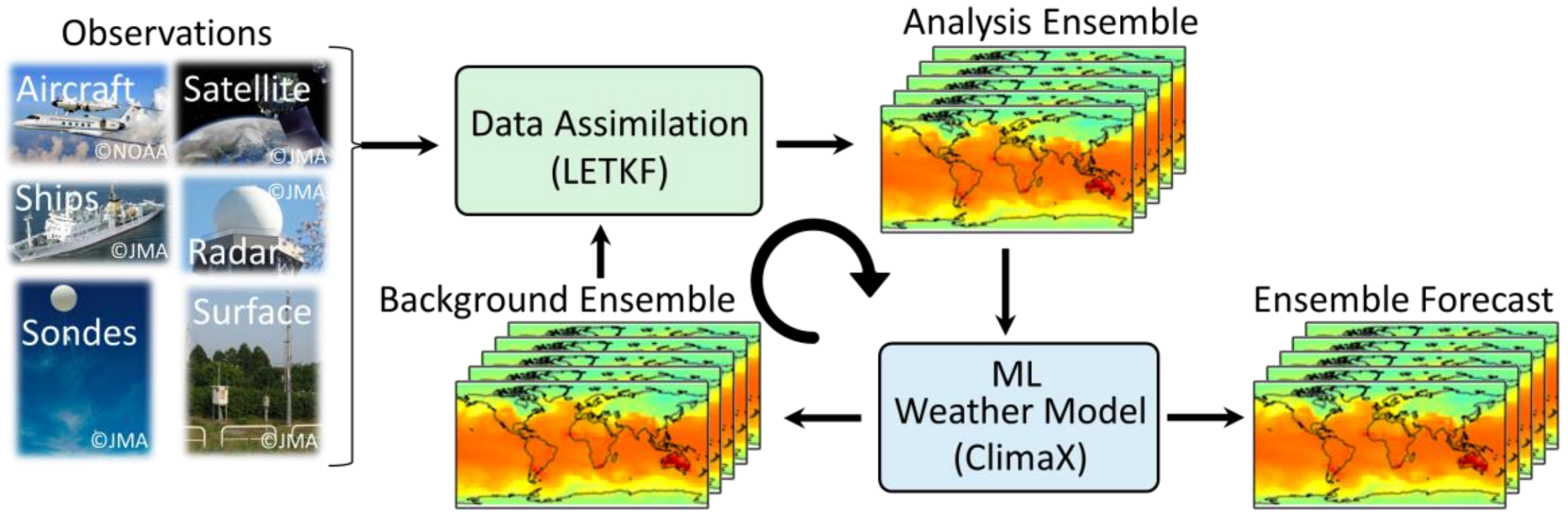

**Figure 1.** The structure of ClimaX-LETKF. The MLWP model ClimaX computes background ensemble from analysis. Observations and background ensemble are integrated by data assimilation to yield analysis ensemble followed by 6-hour ensemble forecast to the next time of data assimilation.

## 2.2. NCEP ADP global observation data

NCEP ADP Global Upper Air and Surface Weather Observations (NCEP 2008; NCEP observation data/dataset hereafter) is a meteorological observation dataset in PREPBUFR format (Keyser 2013), which includes surface observations, radiosonde data, aircraft reports, radar observations, and satellite-derived wind data. The key meteorological variables used in NWP models such as pressure, wind speed, wind direction, temperature, and humidity are provided. It also includes information on location, observation errors, data quality, processing history, and other metadata. Observations are categorized according to the measurement instruments used. An overview of the data from the main categories is shown in Figure 2.

## 2.3. Data assimilation

### 2.3.1. Local Ensemble Transform Kalman Filter

Given its efficiency in parallel computing and performance with a small ensemble size, we use the LETKF (Hunt et al. 2007), a specific formulation of the ensemble Kalman filter (EnKF; Evensen 1994), with 20 ensemble members. The ClimaX-LETKF system was developed based on the SPEEDY-LETKF system by Kotsuki et al. (2025). The LETKF is a widely used DA algorithm in operational weather forecasting systems such as Deutscher Wetterdienst (DWD), Environment and Climate Change Canada (ECCC) and Japan Meteorological Agency (JMA). Like other EnKFs, it tends to produce underdispersive ensembles, primarily due to model imperfections and insufficient ensemble size. To mitigate this, background error covariance or analysis perturbations must be inflated (Whitaker and Hamill 2002). Based on preliminary experiments, we employ two covariance relaxation methods:

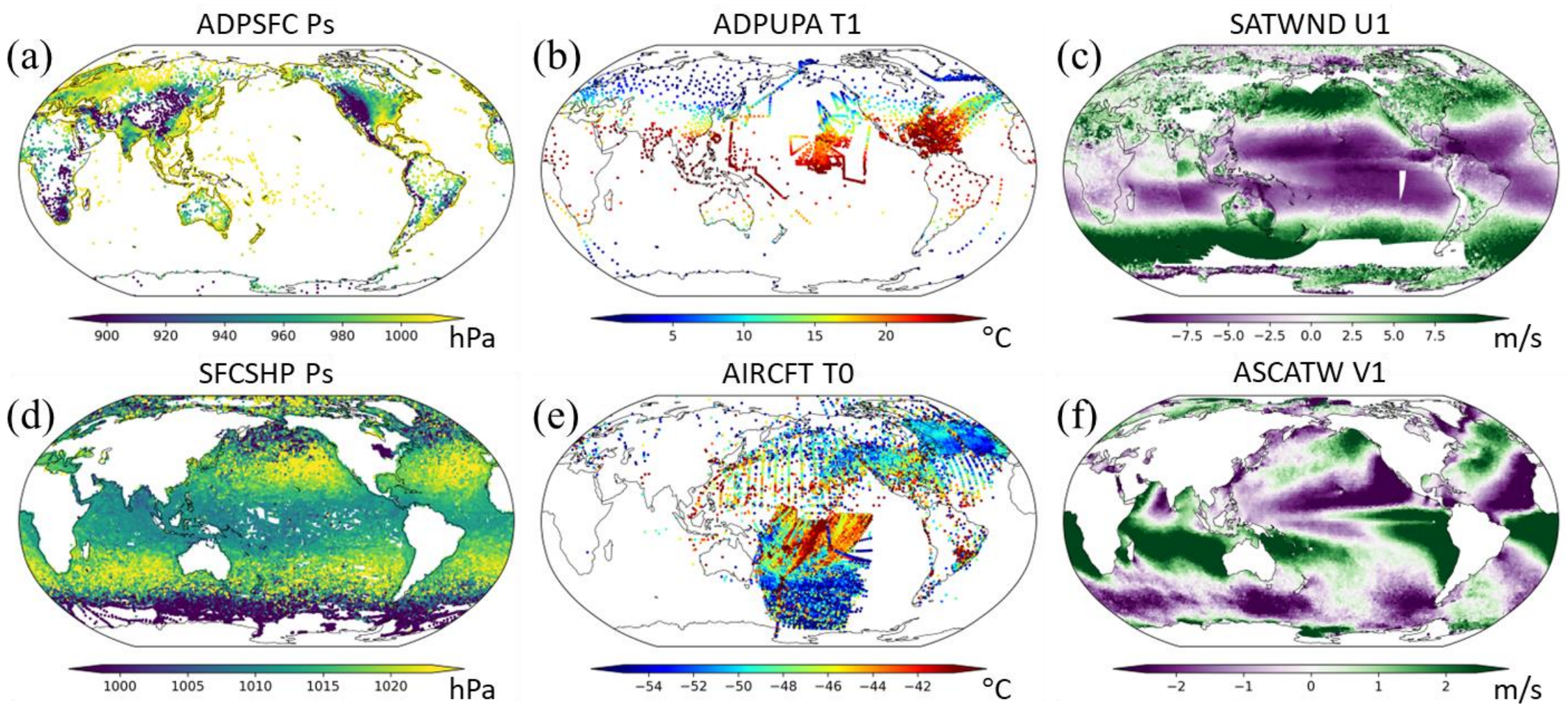

**Figure 2.** An overview of NCEP observation data in 2016. Surface pressure of (a) ADPSFC (surface synoptic reports) and (d) SFCSHP (surface marine reports), temperature of (b) ADPUPA (radiosondes) in the 1st layer and (e) AIRCFT (aircrafts) in all layers, zonal and meridional winds of (c) SATWND (atmospheric motion vectors from geostationary satellites) and (f) ASCATW (ASCAT scatterometer over the ocean) in the 1st layer, respectively. Here the 1st layer is the model layer of 950 hPa, closest one to the surface.

relaxation to prior spread (RTPS; Whitaker and Hamill 2002) and relaxation to prior perturbation (RTPP; Zhang et al. 2004). The RTPS updates the analysis perturbation $\delta\mathbf{x}_i^a$ at each grid point, adjusting its standard deviation according to the following equation:

$$\sigma^a \leftarrow (1-\alpha)\sigma^a + \alpha\sigma^b, \tag{1}$$

where σ represents the ensemble standard deviation, and superscripts $b$ and $a$ indicate the prior (background) and posterior (analysis) states, respectively. This can be rewritten as:

$$\delta\mathbf{x}_i^a \leftarrow \delta\mathbf{x}_i^a \left(\alpha\frac{\sigma^b - \sigma^a}{\sigma^a} + 1\right), \tag{2}$$

where $\delta\mathbf{x}_i \equiv \mathbf{x}_i - \bar{\mathbf{x}}$ denotes the deviation of the $i$th ensemble member $\mathbf{x}_i$ from the ensemble mean $\bar{\mathbf{x}}$. Alternatively, the RTPP updates analysis perturbation by replacing $\delta\mathbf{x}_i^a$ with a weighted mean of $\delta\mathbf{x}_i^a$ and $\delta\mathbf{x}_i^b$ for each ensemble member $i$, as follows:

$$\delta\mathbf{x}_i^a \leftarrow (1-\alpha)\delta\mathbf{x}_i^a + \alpha\delta\mathbf{x}_i^b. \tag{3}$$

The RTPS and RTPP include a tunable relaxation parameter α, which is examined in a subsequent section.

Localization weights $w$ at observation points are calculated by:

$$w = \begin{cases} \exp[-0.5\{(d_h/\rho_h)^2 + (d_v/\rho_v)^2\}] & \text{if } d_h < \sqrt{10/3}\,\rho_h \cap d_v < \sqrt{10/3}\,\rho_v \\ 0 & \text{others} \end{cases}, \quad (4)$$

where $d_h$ is the great-circular distance (km) and $d_v$ is the difference between the natural logarithm of the pressure (log hPa) of the observation point and that of the assimilated grid point. Tunable parameters are set as $\rho_h = 600$ (km) and $\rho_v = 0.1$ (log hPa) based on preliminary investigations (Kotsuki et al. 2025). Model variables are spatially interpolated by observation operators to compute observation departures. A gross error check is applied to all observations, rejecting those for which the absolute difference between the observation and the background is more than 10 times larger than the observation error standard deviation (Terasaki et al. 2015).

### 2.3.2. Assimilated observation data

Zonal and meridional wind components, temperature, specific humidity, and surface pressure are assimilated every 6 hours. Among the land surface observations, only pressure is used, considering other variables' large heterogeneity. The observation data within 30 min before and after the assimilation time are assimilated, and analysis ensemble are used as initial states for subsequent ensemble forecast. If an observation error is missing in the dataset, a predefined constant is assigned to each variable as follows: 1 (m/s) for zonal and meridional winds, 1 (K) for temperature, 0.01 (kg/kg) for specific humidity, and 1 (hPa) for surface pressure. Details of preprocessing including unit conversion and observation thinning are described in Supplement 3.

## 3. Experiments and results

To investigate an appropriate method of covariance inflation and its parameters for the ML-based model ClimaX, we conducted a 2-year DA cycle with 20 ensemble members starting at 00:00 UTC on 1 January, 2016, using two different covariance inflation methods: RTPS and RTPP. Since DA cycle experiments with 20 and 40 ensemble members showed similar results (Supplement Figure S4, S5), the former results are discussed in the manuscript. WeatherBench data at 6-hour intervals from 00:00 UTC on January 1, 2007, were used as the initial conditions for 20 ensemble members. Below, we mainly discuss errors in temperature at 500 hPa because it is one of the most important variables for medium-range weather forecasts. Note that similar results are observed for other variables, such as zonal wind at 700 hPa, meridional wind at 850 hPa, and surface pressure (Supplement Figure S5).

Figure 3 shows the time series of the latitude-weighted root mean square errors (RMSEs) and the latitude-weighted global means of ensemble spreads of background temperature at 500 hPa for experiments using RTPS and RTPP with their respective relaxation parameters. The system remains stable throughout the experimental period with an appropriate covariance inflation method and relaxation parameter (Figure 3a). In the RTPS with a parameter of 1.3, a sharp increase in RMSE is seen around February 2016 and again from June 2016, and similar increases or unstable trends in RMSE are also seen with other relaxation parameters. Since no significant decrease in spread is identified, the main cause is likely not an underdispersion but rather a physical imbalance. Furthermore, for all relaxation parameters, increased RMSE and declined spread continues after April 2017. With RTPP, the differences in the RMSEs among experiments with different relaxation parameters are subtle and the best performance is achieved with a parameter of 0.90, where the ensemble spread is closest to the RMSE (Figure 3b). Such behaviors differ from those of NWP models in terms of both the relaxation methods and the relaxation parameters. In general, NWP models tend to be more stable with RTPS than with RTPP (e.g. Kotsuki et al. 2017; Whitaker and Hamill 2012). In contrast, RTPP consistently shows lower and more stable RMSEs across almost all periods in our ClimaX-LETKF system. The system also requires higher RTPP inflations than NWP model-based systems (e.g., 0.80 in Kotsuki et al. 2017 and 0.76 in Whitaker and Hamill 2012), indicating the less-chaotic nature and higher imperfection of the model we trained (Anderson and Anderson 1999; Evensen et al. 2022; Whitaker and Hamill 2002).

Figure 4 shows the spatial distributions of the background RMSEs of several important variables, temperature at 500 hPa, zonal wind at 700 hPa, meridional wind at 850 hPa and surface pressure, as well as the background ensemble spreads and the number of observations in the grids, both averaged over the experimental period except for the first month (i.e., from February 2016 to December 2017) considering the elimination of spin-up periods. The relation between spread and RMSE has a similar spatial pattern among these four variables. Spreads tend to be excessive in the low- to mid-latitude area, Greenland and the Antarctica and too large over oceans in the high-latitude area compared to RMSEs. The underdispersive tendency over land is clearer in the lower layers (see also Supplement Figure S6). Negative correlation was found between the errors and the number of observations in some variables; errors tend to be larger in sparsely observed regions. Peason correlation coefficient between RMSE and the number of observations were -0.31, -0.45, -0.40 and -0.15 for temperature, zonal wind, meridional wind and surface pressure in correspondent layers, respectively. Temperature errors are small in low-latitude regions and relatively low over land in the Northern Hemisphere, where observations are relatively dense, but significantly large in high-latitude regions of the Southern Hemisphere, where observations are very sparse. Note that modest temporal changes in low-latitude regions may also have substantial effects on these results. Wind errors exhibit similar patterns to temperature, reflecting the impact of observations in reducing errors. However, cases such as temperature over Africa and winds over Tibetan Plateau, where RMSE and spreads are small despite few observations, can also be found; these are mainly considered to be due to the inherent ease of prediction in those areas. In contrast, surface pressure errors are significantly smaller over low-latitude oceans despite sparse observations compared to land, reflecting no orographic impact there and small temporal changes in low-latitude regions. Large errors in surface pressure observed over extensive areas of Russia and northern Canada despite the presence of dense observations may be due to relatively large discrepancies between the model's topography and the real-world terrain in these regions.

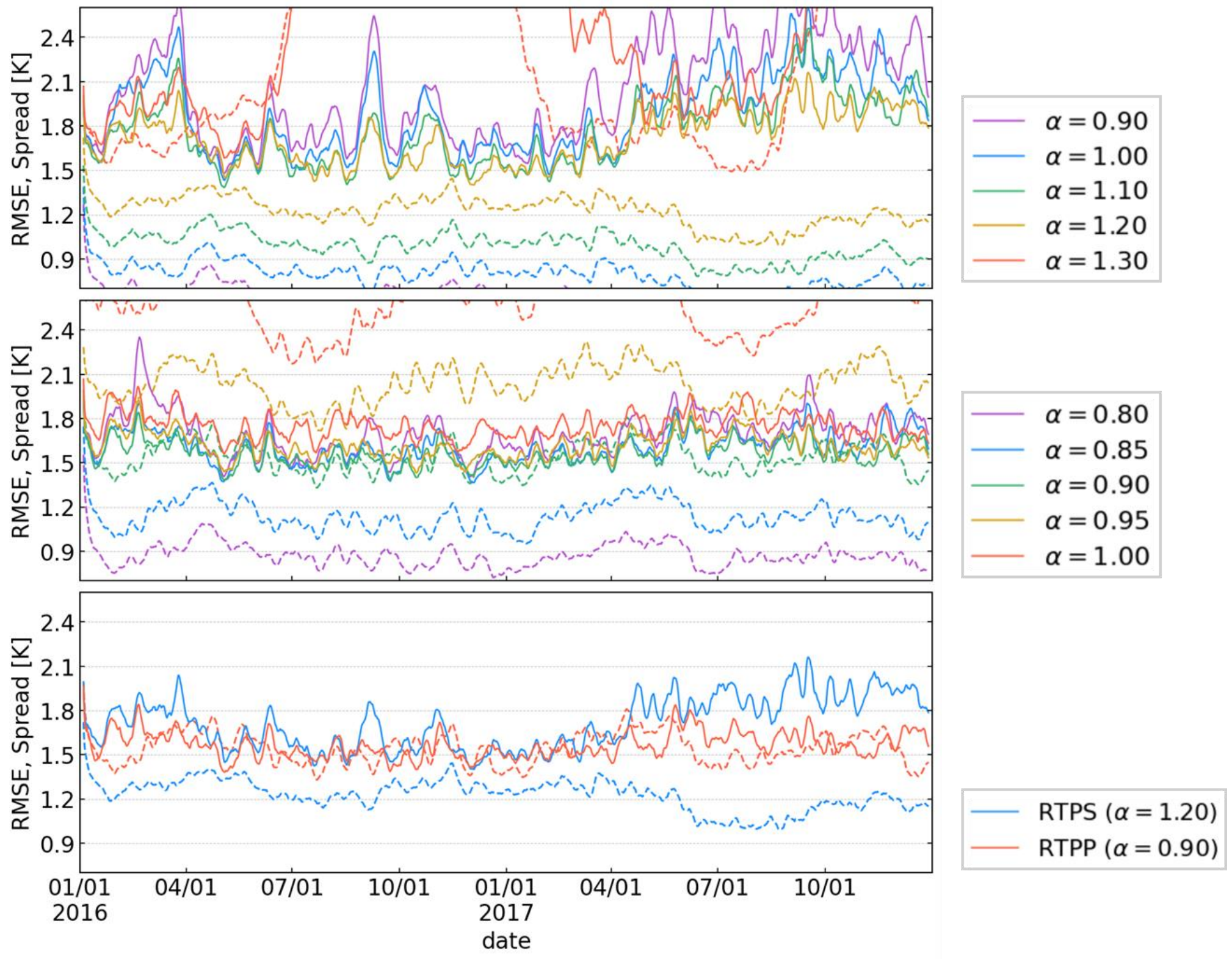

**Figure 3.** Time series of latitude-weighted global means of RMSEs verified against WeatherBench (solid lines) and ensemble spreads (dashed lines) of background temperature at 500 hPa for the data assimilation experiments. (a) and (b) show the results of different parameters for RTPS and RTPP, respectively. (c) compares the RMSEs and ensemble spreads of RTPS and RTPP with their best-performing parameters $\alpha$.

To analyze the potential ability of this ensemble forecasting system to represent its own uncertainties in the forecasts, we investigate changes in RMSE and ensemble spread with respect to forecast lead time. We employ 5-day ensemble forecasts using RTPS and RTPP, with their optimal parameters, for January 2017. The global-mean RMSE and ensemble spread, averaged over 124 initial times, are shown in Figure 5. Ensemble spread indicates the uncertainty of prediction, and it is desirable that the spread is comparable to the RMSE. However, the increase in the spread is more gradual than that of the RMSE, resulting in a larger gap between them as the lead time increases. This reflects the less-chaotic nature of the data-driven model we used relative to physics-based NWP models, which is also consistent with the interpretation of the DA cycle experiments. The spread shows different behaviors between the two covariance inflation methods. In the RTPP experiment, the spread increases with an increase in lead time. With RTPS, the spread initially decreases at the shortest lead time of 6 h but then increases over longer lead times. As a result, RTPP consistently produces a larger spread that remains closer to the RMSE than RTPS across all lead times.

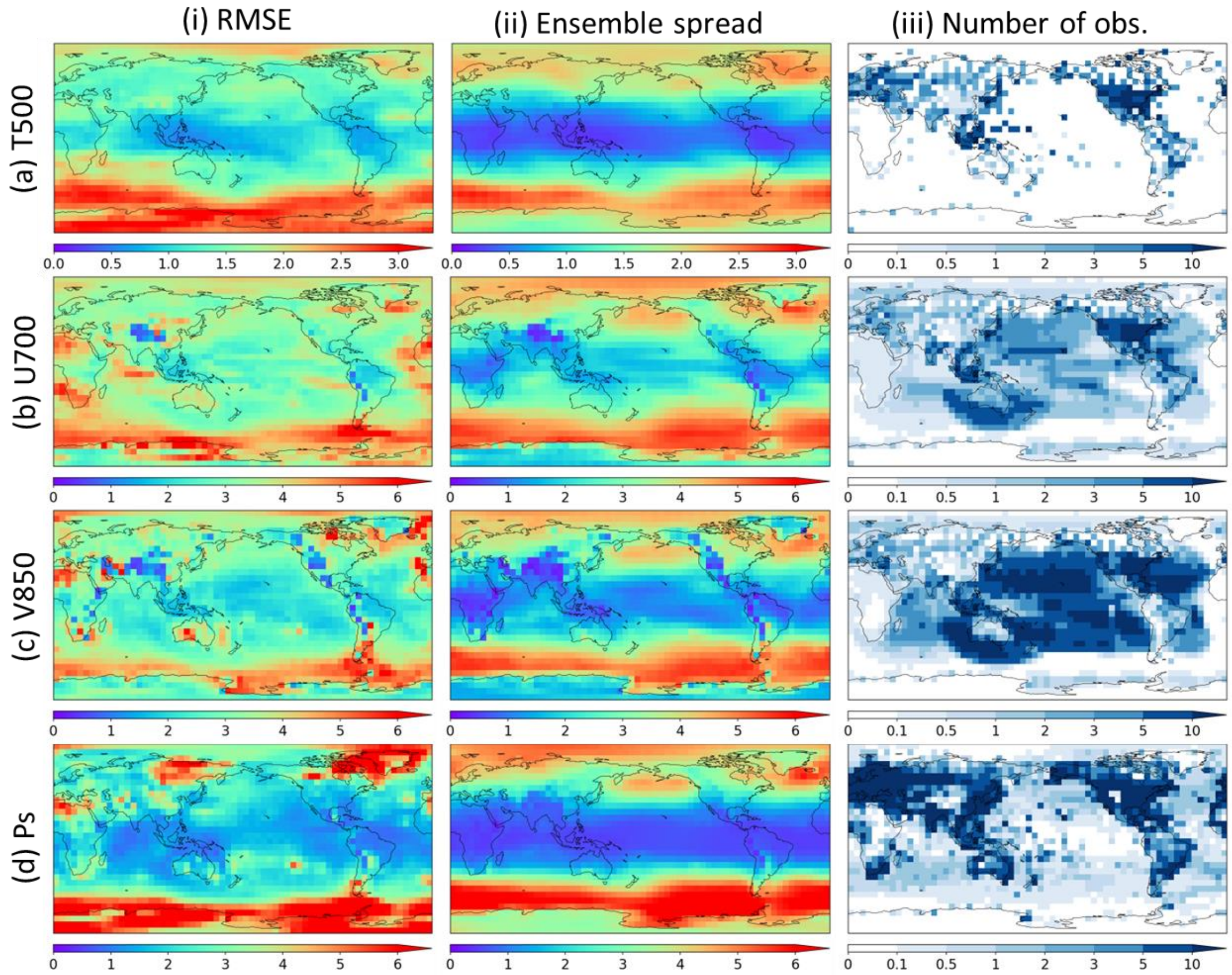

**Figure 4.** Spatial distributions of (i) background RMSE against WeatherBench and (ii) ensemble spread for (a) temperature at 500 hPa (K), (b) zonal wind at 700 hPa (m/s), (c) meridional wind at 850 hPa (m/s) and (d) surface pressure (hPa). The figures in (iii) the rightmost column are the number of observations of corresponding variables after thinning in each grid at the corresponding layers (i.e., 500 hPa, 700 hPa, 850 hPa, and surface level, respectively). Ensemble spread and the number of observations are averaged over February 2016 – December 2017.

## 4. Discussion and Summary

In general, assimilating real observations is more challenging than assimilating synthetic observations, which typically have simplified spatial distributions and error characteristics (e.g. Huang et al. 2024 or Xiao et al. 2024). Another challenge is the spatial representativeness of observations. Real observations represent atmospheric conditions in the limited surrounding area much more localized than model grid sizes, thus they tend to have larger difference from the reference grid-point value than synthetic observations generated by adding noises to the grid-point values. Due to these challenges, even models that perform well in observation system simulation experiments may fail when assimilating real observations.

One well-known issue with MLWP models is their failures to account for the physical balance of the atmosphere. It has been shown that these models fail to capture inter-variable relationships, represent the propagation of perturbations, or account for the rapid growth of small errors in the initial conditions (Selz and Craig 2022; Slivinski et al. 2025), indicating that they do not adequately represent the physical laws governing the atmosphere in contrast to NWP models. These findings suggest that MLWP models are less capable of constraining the fields to the attractor of the atmosphere than NWP models. The ClimaX model we trained also has this problem (Kotsuki et al. 2025). In general, for medium-range weather predictions, NWP models tend to be more stable and to produce lower errors with RTPS rather than RTPP (e.g., Kotsuki et al. 2017; Whitaker and Hamill 2012). However, this did not hold in our study, which is likely to be due to the inherent characteristics of our MLWP model. Since physical balances are not enforced during DA, they may exhibit undesirable features, such as spatial discontinuities or physical imbalances – in other words, a divergence from the attractor of the atmosphere – which can lead to further divergence in subsequent timesteps. As a result, MLWP models may struggle to predict fields with the same accuracy as during validation with reanalysis data, let alone restore physical balances. The RTPP method generates an analysis ensemble where the perturbation is blended with the background perturbation. In this study, the lowest RMSE was achieved with a relaxation factor of 0.9, indicating that the influence of the LETKF-based analysis perturbations was reduced, while the perturbations from the ensemble forecasts played a dominant role. In contrast, the RTPS method multiplicatively inflates the LETKF-based analysis perturbations, causing the imbalances in the analysis to strongly impact subsequent forecasts. This difference is considered the main factor behind the variation in stability and accuracy between the two covariance inflation methods.

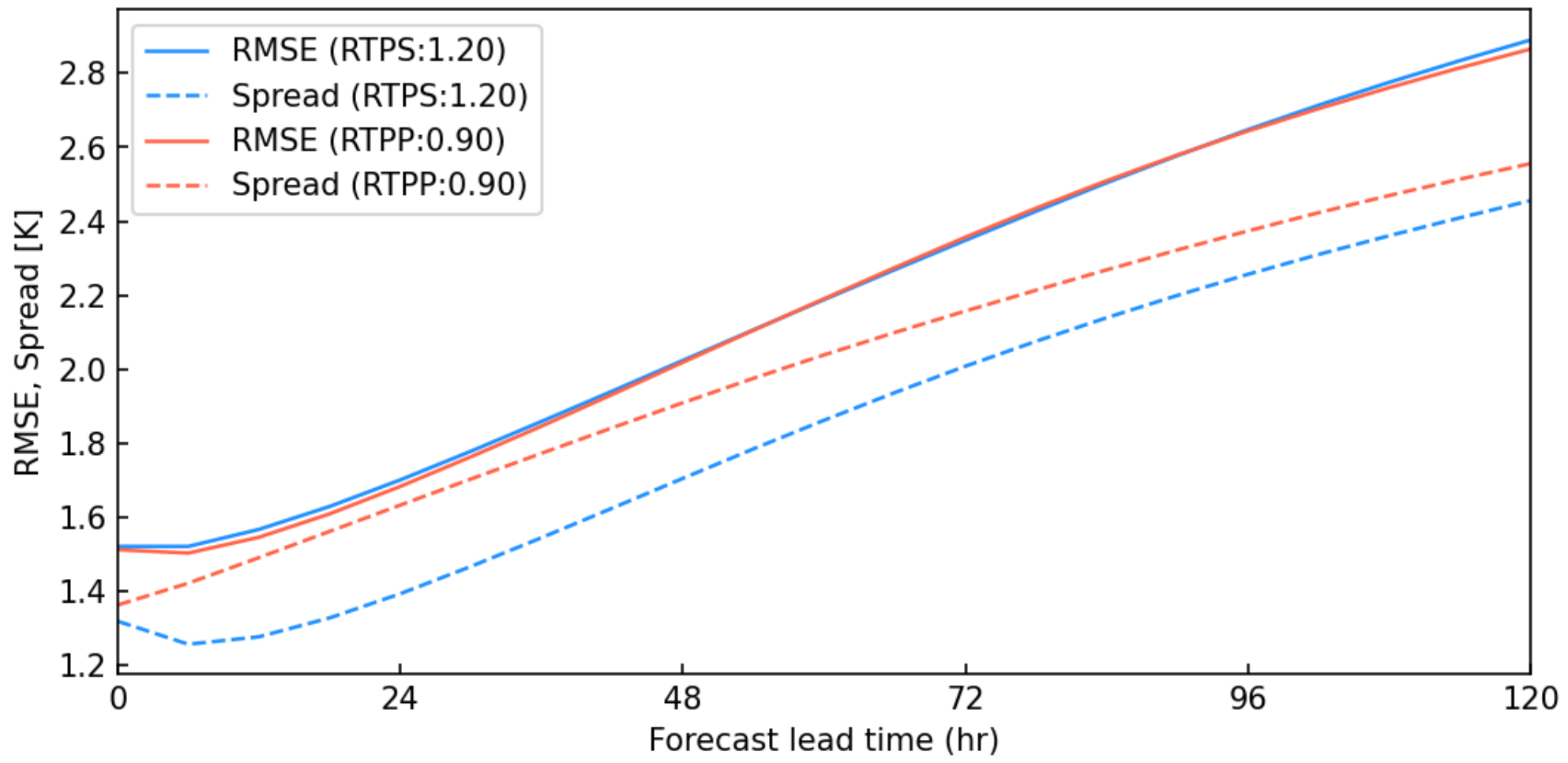

**Figure 5.** Latitude-weighted RMSE against WeatherBench and the latitude-weighted global mean of ensemble spread of temperature at 500 hPa (K) as a function of forecast lead time with the initial times from the beginning to the end of January 2017. The horizontal axis shows forecast lead time (hour) and the vertical axis shows RMSE and ensemble spread (K). Solid lines and dashed lines indicate RMSE and ensemble spread, respectively, and blue and red color indicate RTPS and RTPP, respectively.

Although there is still much room to improve the accuracy of ClimaX-LETKF, our successful demonstration of multi-year ensemble weather forecasting using an MLWP model that assimilates real observations provides valuable insights for developing more stable and accurate ML-based prediction systems, independently of NWP models, bringing us closer to realizing weather forecasting by MLWP models.

**Acknowledgements**

The authors declare that they have no conflicts of interest. This study was partly supported by the JST Moonshot R&D (JPMJMS2389), the Japan Society for the Promotion of Science (JSPS) KAKENHI grants JP21H04571, JP21H05002, JP22K18821, JP 25H00752, the Japan Aerospace Exploration Agency (JAXA) Precipitation Measuring Mission (PMM) (grant no. JX-PSPC-539791), and the IAAR Research Support Program and VL Program of Chiba University.

**Supplementary Materials for**
**Bridging Artificial Intelligence and Data Assimilation:**
**The Data-driven Ensemble Forecasting System ClimaX-LETKF**

## S1. Definition of RMSE and ensemble spread

Root mean square error (RMSE) evaluates global errors between ensemble mean of background or analysis and reference field. Latitude-weighted RMSE at a certain time is defined by:

$$RMSE = \sqrt{\left\{\sum_{i=1}^{n} A_i(\bar{x}_i - x_i^{true})^2\right\} \Big/ \left(\frac{1}{n}\sum_{i=1}^{n} A_i\right)}, \quad (S1)$$

where subscript $i$ is a grid index, $\bar{x}_i$ is an ensemble mean of background or analysis field, $x_i^{true}$ is a reference field, $A_i$ is an area of $i$th grid, and $n$ is the number of grids.

Ensemble spread evaluates the potential ability of the system to represent its own uncertainty. Its latitude-weighted-mean at a certain time is defined by:

$$Spread = \sqrt{\left\{\sum_{i=1}^{n} A_i\left(\frac{1}{l-1}\sum_{k=1}^{l}\left(x_i^{(k)} - x_i^{true}\right)^2\right)\right\} \Big/ \left(\frac{1}{n}\sum_{i=1}^{n} A_i\right)}, \quad (S2)$$

where $x_i^{(k)}$ is the $k$th ensemble member of background or analysis and $l$ is the ensemble size.

## S2. Preprocessing of PREPBUFR data

S2-1. Unit conversion

Virtual temperature $T_v$ (K) is converted to real temperature $T$ (K) with the following equation:

$$T = \frac{T_v}{1 + \left(\frac{1}{\varepsilon} - 1\right)Q}, \quad (S3)$$

where $\varepsilon$ is the ratio of molecular weight of water vapor to that of dry air and constant at 0.622, and $Q$ (kg/kg) is specific humidity.

Errors of humidity are recorded in terms of relative humidity, not specific humidity. An error of relative humidity $RH_{err}$ (%) is converted to an error of specific humidity $Q_{err}$ (kg/kg) with the following equations:

$$P_w = P\frac{Q}{\varepsilon + (1-\varepsilon)Q}, \quad (S4)$$

$$P_{ws} = 6.1121 \exp\left[\left(18.678 - \frac{T_c}{234.5}\right)\frac{T_c}{257.14 + T_c}\right], \quad (S5)$$

$$P_{w_{err}} = P_{ws}\frac{RH_{err}}{100}, \quad (S6)$$

$$Q^{\pm} = \frac{\varepsilon\left(P_w \pm P_{w_{err}}\right)}{P - (1-\varepsilon)\left(P_w \pm P_{w_{err}}\right)}, \quad (S7)$$

$$Q_{err} = (Q^{+} - Q^{-})/2, \quad (S8)$$

where $P_{ws}$ (hPa) is saturated water vapor pressure calculated by the Buck equation (Buck 1981), $T_c$ (°C) is temperature, $P_w$ (hPa) is water vapor pressure and $P_{w_{err}}$ (hPa) is its error, $Q^{+}$ (kg/kg) and $Q^{-}$ (kg/kg) are specific humidities when water vapor pressure is $P_w + P_{w_{err}}$ and $P_w - P_{w_{err}}$, respectively. In general, $Q^{+}$and $Q^{-}$ are assymmetric to $Q$ because function $Q = f(P_w)$ is non-linear, but whether error is positive or negative is not distinguished in assimilation, so $Q_{err}$ is approximated as stated above.

S2-2. Observation Thinning

Observation thinning selects observations used in data assimilation from NCEP observation data. The algorithm is designed with consideration of these two points:

- Observations in sparsely observed areas are prioritized.
- When removing excess observations, the remaining observations are distributed as homogeneously as possible. (Figure S1)

To evaluate the importance of observation points, three concepts are introduced: weight of observation point, observation density, and contribution of observation point.

Firstly, the larger the weight of observation point is, the more likely to be selected this observation point is. Weight $w_{i,j}$ of the $i$th observation point $obs_i$ for the $j$th grid point $grid_j$ is defined as follows:

$$w_{i,j} = w_{i,j}^{d} w_{i}^{q}, \quad (S9)$$

$$w_{i,j}^{d} = \begin{cases} \exp[-0.5\{(d_h/r_h)^2 + (d_v/r_v)^2\}] & \text{if } d_h < \sqrt{10/3}\, r_h \cap d_v < \sqrt{10/3}\, r_v \\ 0 & \text{else} \end{cases}, \quad (S10)$$

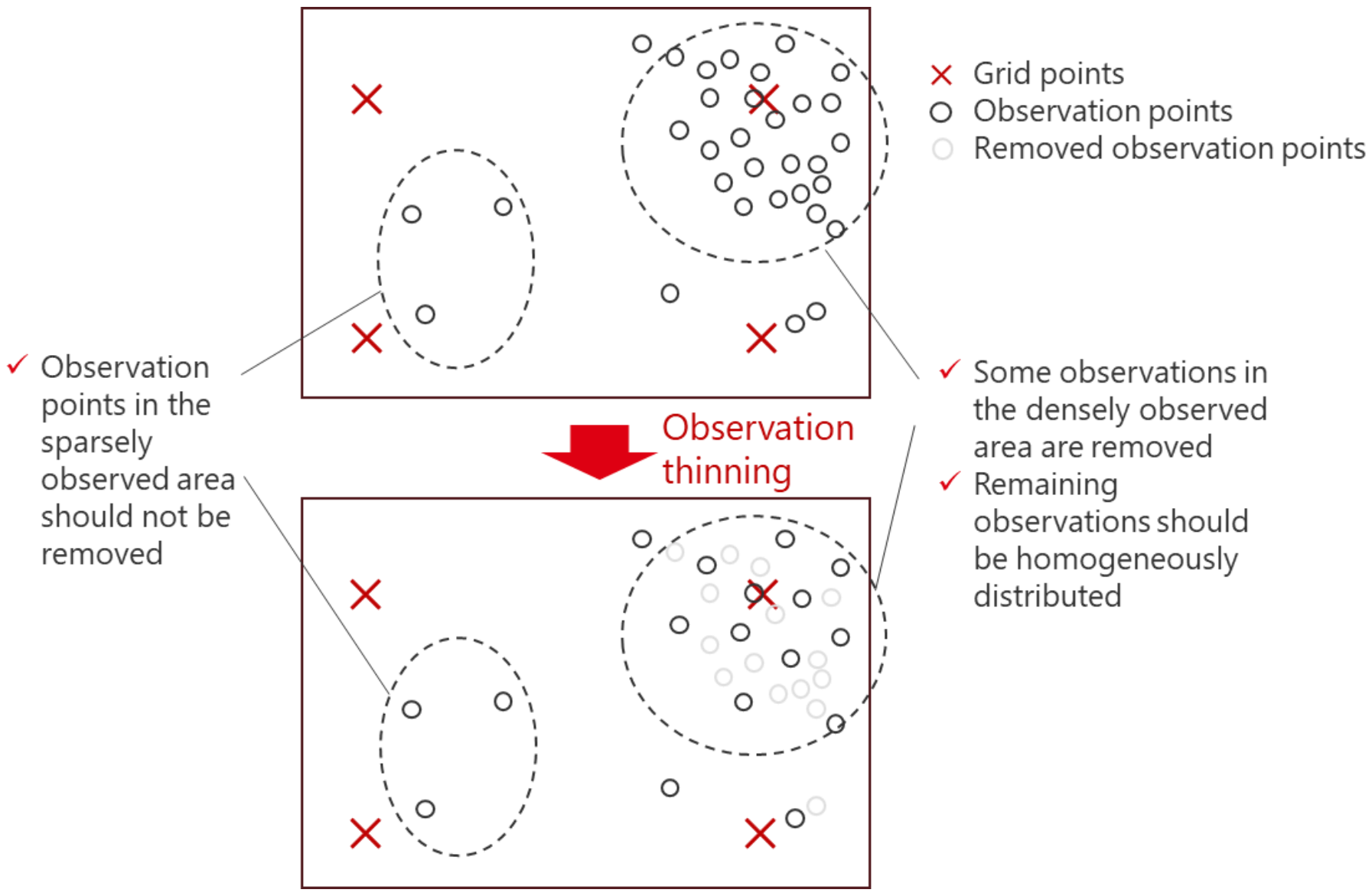

**Figure S1.** Schematics of observation thinning. In general, the distribution of observation points is highly inhomogeneous. When removing observation points, those in sparsely observed areas should not be removed, and those in densely observed areas should be removed so that remaining ones are distributed as homogeneously as possible.

where $w^d$ and $w^q$ are distance weight and quality weight, respectively. $w^d$, which is calculated by the same formula to that of the localization weight of LETKF, is in the range of $[0,1]$. $d_h$ is the great-circular distance on the Earth's surface and $d_v$ is the absolute difference of natural logarithm of pressure between $obs_i$ and $grid_j$. $r_h$ and $r_v$ are tunable parameters, respectively, and $r_h = 500$ (km) and $r_v = 0.1$ (log hPa) in this study. The Earth is approximated as a perfect sphere of radius approximately 6371.0072 km, a sphere of the same surface. In this manuscript, "local" means that two points are inside the localization cut-off radius distance of each other. Therefore, "local observation points of $i$th grid point" is defined as a set of observations such that $w_{i,j} > 0$. "Local grid points of $j$th observation point" is defined in the same manner.

This concept is applied to the relationship between two observation points. To avoid concentration of the selected observation points, those around the selected observation point are less prioritized. When $obs_j$ is selected, weights of its local observation points are updated by:

$$w_{i,j'} \leftarrow w_{i,j'} \left(1 - {w^d_{j,j'}}^2\right), \tag{S11}$$

where $j'$ is the global index of the local observation point of $obs_j$.

Quality weight $w^q$, which also can be in the range of $[0,1]$, corresponds to the value of the "quality marker" (qmk hereafter) of observations. This value takes an integer of 0 to 15, or the missing value. Since using observations with qmk > 3 is not recommended, in this study, weights of such observations are basically set to zero, meaning that they are definitely not selected in the thinning process. For observations with qmk of 0, 1, 2, and 3, $w^q$ of 1.0, 0.8, 0.4, and 0.1 are applied, respectively. If qmk is missing, 0.4 is given to $w^q$.

Second, observation density is defined for each grid point $grid_i$ as a summation of the weights of its local observations. There are two kinds of observation density: density of all observations $W_i^{all}$ and density of the selected observations $\widetilde{W}_i$. From their definitions, $W_i^{all}$ is constant but $\widetilde{W}_i$ increases if a local observation point is newly selected; $\widetilde{W}_i$ is initially 0, and if all local observations are selected, this is equal to $W_i^{all}$.

Finally, contribution $c_{j,i}$ indicates the impact of selection of the $j$th observation point on the observation density of the $i$th grid, defined as follows:

$$c_{j,i} = \max\left(0, w_{i,j}\frac{W_i^{max} - \widetilde{W}_i}{\widetilde{W}_i + \varepsilon}\right), \tag{S12}$$

where $W_i^{max}$ is a threshold of observation density and $\varepsilon$ is a tiny constant value 1E-10 for enabling calculation while $\widetilde{W}_i$ is zero. The observation point with the highest contribution is selected in every iteration step. The smaller $\widetilde{W}_i$ is, and the larger $w_{i,j}$ is, the larger $c_{j,i}$ is. Selecting too many observations is regarded to be ineffective for improving an accuracy of analysis, and this is controlled by the threshold $W_i^{max}$. When $\widetilde{W}_i$ is equal to or greater than $W_i^{max}$, the contribution $c_{j,i}$ of selecting new observations around this grid is zero. $W_i^{max}$ is constant at 1.0 in this study.

Here are the details of the algorithm of observation thinning (partially simplified). $n$ is the number of grid points and $m$ is the number of observations. Thresholds $m^{thresh}$ and $c^{thresh}$ are described below.

---

DO $i = 1, n$

  Search for the local observation points of $grid_i$

  DO $j' = 1, m_i$ ($m_i$ is the number of local observation points of $grid_i$)

    $j$ is the global index of the $j'$th local observation point of $grid_i$.

    Compute $w_{i,j}$

END DO

    Compute $W_i^{all}$

END DO

Set up the update-flags for all observation points.

$m$ is the number of selected observation points.

$m = 0$

DO while $m$ is below the threshold

    DO $j = 1,\ m$

        IF update-flag is not set up, CYCLE

        Fold the update-flag

        DO $i' = 1, n_j$ ($n_j$ is the number of local grid points of $obs_j$)

            $i$ is the index of $i'$th local grid point of $\mathrm{obs}_j$.

            Compute $c_{j,i}$.

        END DO

        Find the highest contribution $c_j^{\max} = \max_i c_{j,i}^{max}$.

    END DO

    $j^{\mathrm{add}} = \operatorname{argmax}_j c_j^{max}$

    IF $c_{j^{add}}^{max}$ is below the threshold, EXIT

    Add $obs_{j^{add}}$ to the set of assimilated observation points.

    DO $i = 1,\ n$

        Update $\widetilde{W}_i$.

    END DO

    IF $\max(\widetilde{W}_i)$ is above the threshold, EXIT

    $m\ =\ m + 1$

    DO $i' = 1,\ n_{j^{add}}$

        $i$ is the index of the $i'$th local grid point of $obs_{j^{add}}$.

        DO $j' = 1,\ m_i$

            $j$ is the index of the $j'$th local observation point of $grid_i$

            IF $obs_j$ is not local for $obs_{j^{add}}$, CYCLE

            Set up the update-flag for $obs_j$

Update $w_{i,j}$, the weight of $obs_j$: $w_{i,j} \leftarrow w_{i,j}\left(1 - {w^d_{j,j^{\mathrm{add}}}}^2\right)$

END DO

END DO

END DO

---

In this algorithm, two thresholds are defined to stop iteration: one is the contribution $c^{thresh}$, and the other is the maximum number of selected observation points $m^{thresh}$. As observation points selected, the weights of surrounding observation points get smaller and observation densities of the grid points get larger, leading to smaller value of $c^{max}_{jadd}$. When this is small enough, namely less than $c^{thresh}$, selection is stopped. $m^{thresh}$ is used for explicitly limiting the number of selected observation points.

Figure S2 illustrates the changes in distributions of observations after thinning in two example regions. In areas with a high density of observations, redundant observation points are removed, and the remaining points are generally distributed uniformly. In contrast, in sparsely observed areas, observations are retained without thinning.

## S3. Initialization method of ClimaX

We modified the initialization method of the convolutional input-encoding layer (referred to as "variable tokenization" in ClimaX) based on preliminary experiments. In the original code of the ClimaX, parameters in the convolutional layer are initialized using a truncated normal distribution with a standard deviation of 0.02. We removed this initialization scheme and instead adopted the default initialization method of PyTorch, a machine learning library, which applies Kaiming Uniform Initialization (He et al. 2015), a weight initialization method for neural networks with ReLU-family activations that samples weights from a zero-mean uniform distribution scaled so that the output variance does not shrink or explode across layers. With this modification, the model showed better prediction skills in validation (Figure S3) and an improved and more stable performance in data assimilation cycle experiments (results not shown) compared to the original method. However, the model to which the modified initialization method was applied also cannot run even for one year without data assimilation, so there is no inherent difference from the model to which original initialization method was applied.

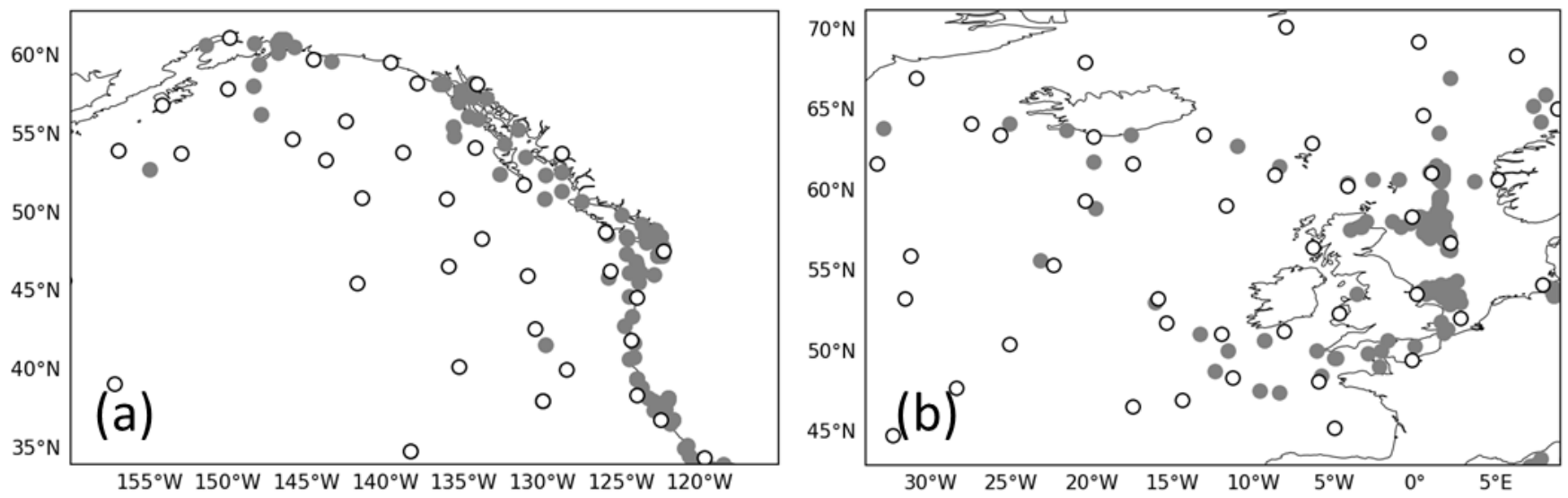

**Figure S2.** Examples of observation thinning. Surface pressure observations by the sea surface observation SFCSHP in (a) western coast of North America and (b) around England. White circles and gray circles indicate observations remain and removed by thinning, respectively.

Anomaly correction coefficient is defined by:

$$ACC = \frac{\sum_{j=1}^{n} L_j \sum_{i=1}^{m} \left(\bar{x}_{i,j}^{b} - x_{i,j}^{clim}\right)\left(x_{i,j}^{true} - x_{i,j}^{clim}\right)}{\sqrt{\sum_{j=1}^{n} L_j \sum_{i=1}^{m} \left(\bar{x}_{i,j}^{b} - x_{i,j}^{clim}\right)^2} \sqrt{\sum_{j=1}^{n} L_j \sum_{i=1}^{m} \left(x_{i,j}^{true} - x_{i,j}^{clim}\right)^2}}, \qquad (S13)$$

where $\bar{x}^b$ is an ensemble mean of background, $x^{true}$ is a reference field of WeatherBench, $x^{clim}$ is a climatology of WeatherBench on 2016, the validated period. $i$ and $j$ are grid indices in longitude and latitude directions, respectively, and $m$ and $n$ are total number of columns and rows of the grid,

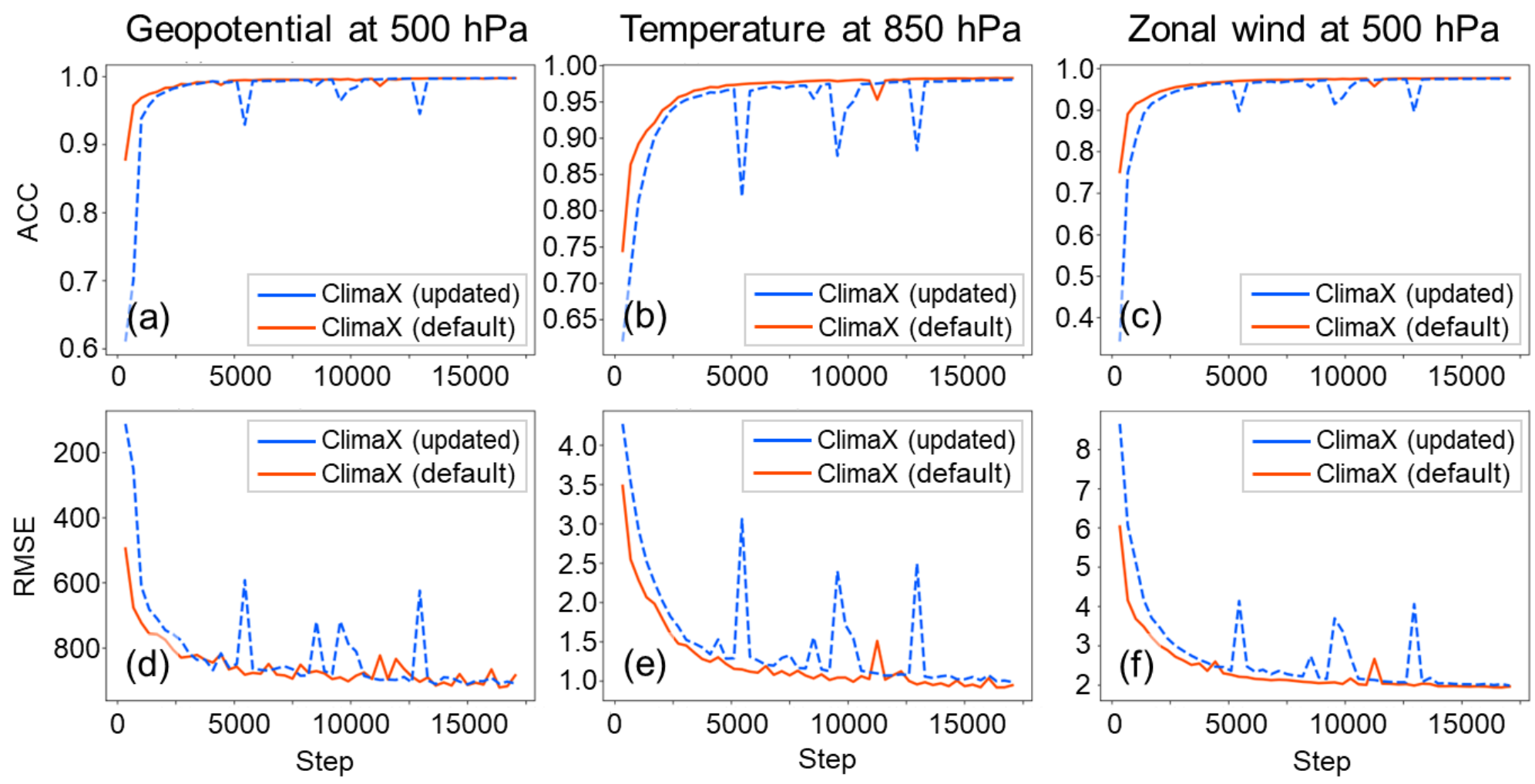

**Figure S3.** Prediction skills for some variables in validation. (a), (b), and (c) shows the latitude-weighted anomaly correlation coefficient (ACC) and (d), (e), and (f) shows the latitude-weighted RMSE. Horizontal axes are steps of learning. Red lines and blue lines show the results of updated and default initialization method, respectively.

respectively. $L_j$ is the latitude weighting factor for the $i$th row of the grid defined as:

$$L_j = \frac{\cos\phi_j}{\frac{1}{n}\Sigma_{j'=1}^{n}\cos\phi_{j'}}, \quad (S14)$$

where $\phi_j$ is the latitude of the $j$th row.

## S4. Results of DA cycle experiments with 40 ensemble members

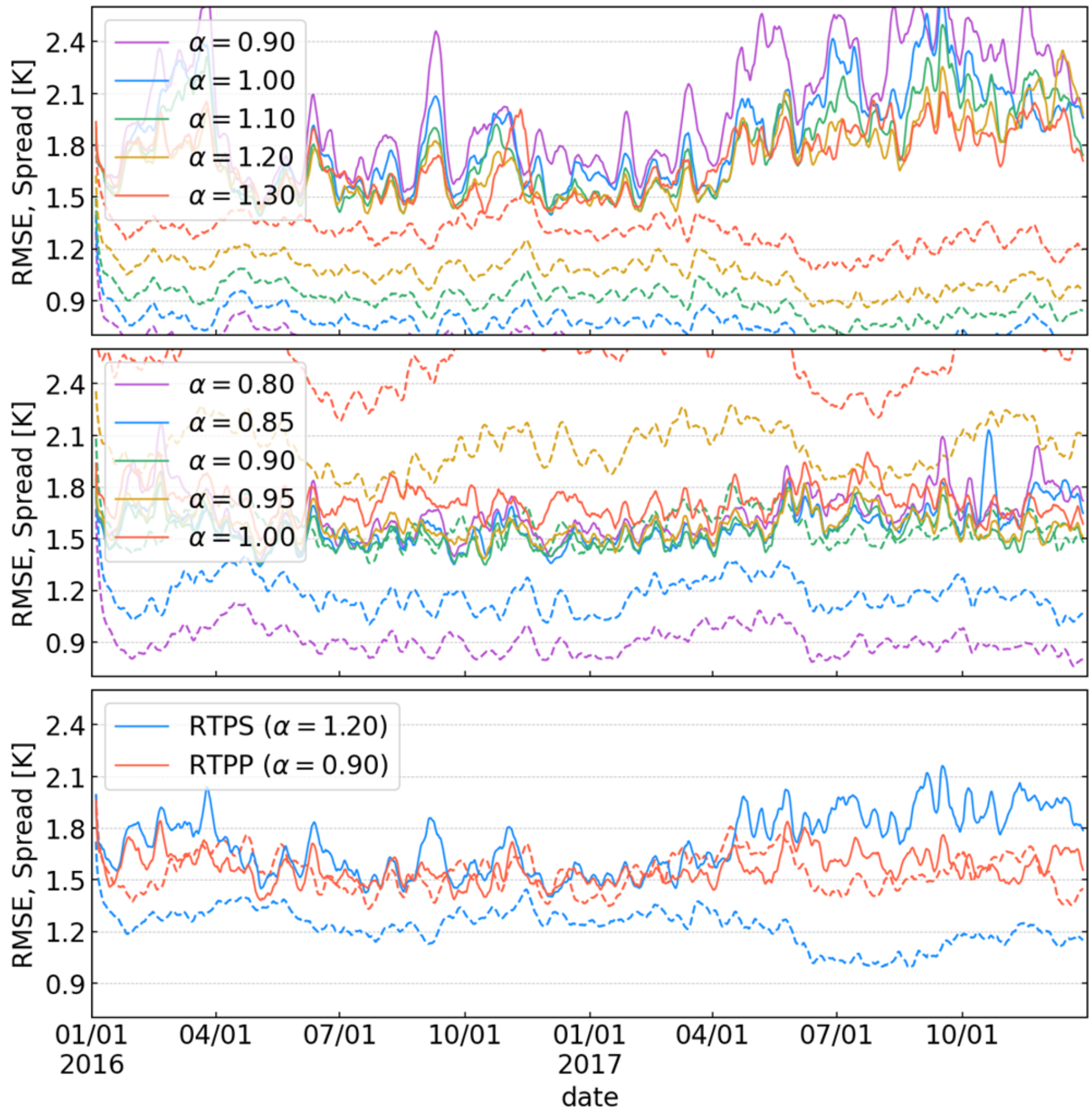

**Figure S4.** Time series of latitude-weighted global means of RMSEs verified against WeatherBench (solid lines) and ensemble spreads (dashed lines) of background temperature at 500 hPa for the data assimilation experiments with 40 ensemble members. (a) and (b) show the results of different parameters for RTPS and RTPP, respectively. (c) compares the RMSEs and ensemble spreads of RTPS and RTPP with their best-performing parameters $\alpha$.

## S5. Experimental results of important variables other than temperature at 500 hPa

Other variables such as surface pressure, geopotential, wind components show the stable results similar to the temperature. Figure S5 shows the timeseries of the latitude-weighted RMSEs and latitude-weighted-mean of the ensemble spreads of these variables. Data assimilation cycle experiments with both 20 members and 40 members exhibited the lower and stable RMSE and ensemble spread closer to the RMSE with RTPP than with RTPS. Underdispersive trends are more pronounced in 40-member experiments.

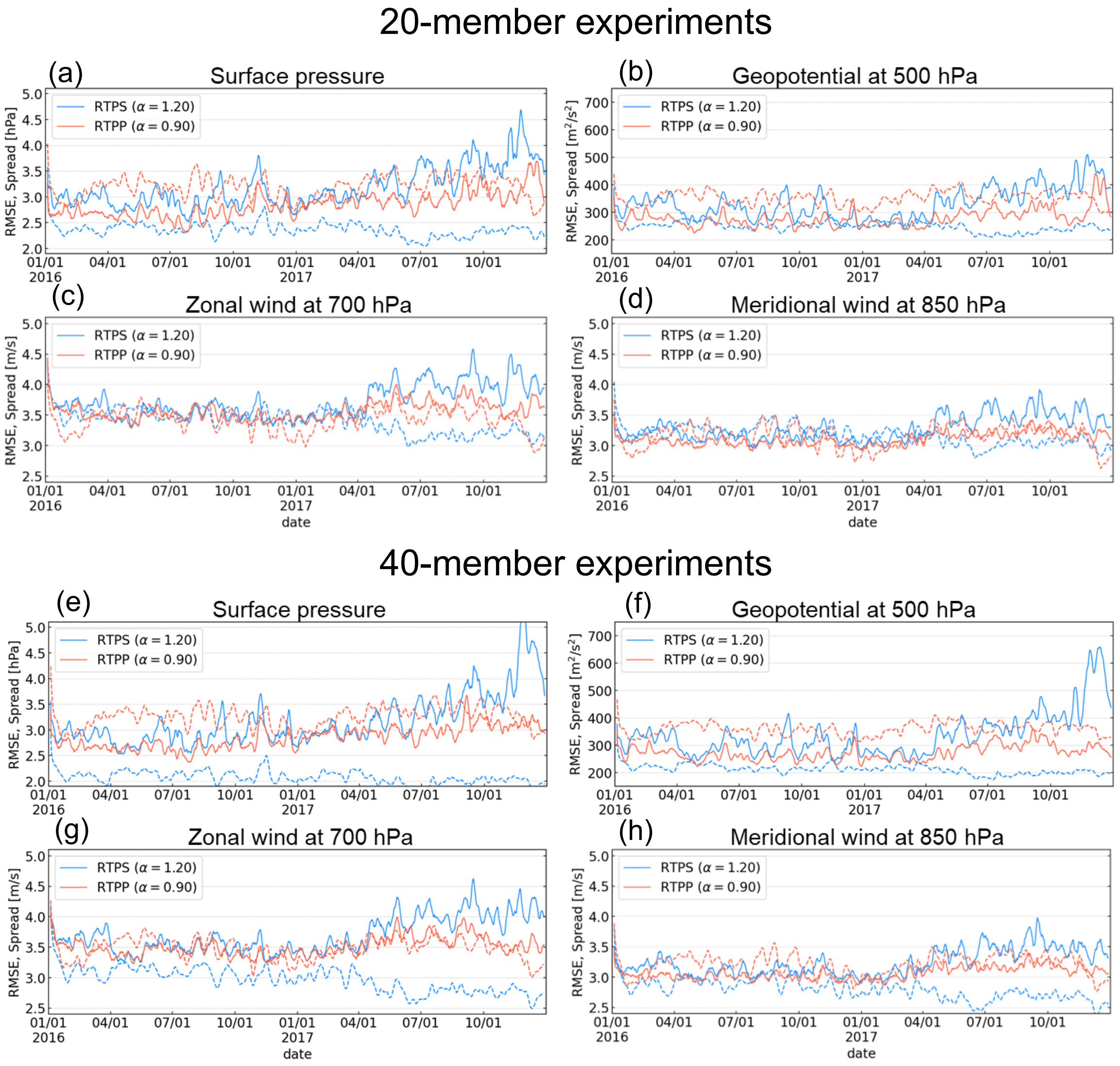

**Figure S5**. Timeseries of RMSEs and ensemble spreads of the backgrounds. (a) – (d) for 20-member experiments and (e) – (h) for 40 member experiments. (a) and (e) surface pressure (hPa), (b) and (f) geopotential at 500 hPa ($m^2/s^2$), (c) and (g) zonal wind at 700 hPa (m/s), and (d) and (h) meridional wind at 850 hPa (m/s). Solid and dashed lines indicate latitude-weighted RMSE and latitude-weighted-mean of ensemble spread, respectively. Blue and red lines indicate the results with RTPS and RTPP with the best parameters for each, respectively. Note that relaxation factor of 1.20 for RTPS and 0.90 for RTPP are the best not only for temperature at 500 hPa but for these variables.

## S6. Spatial distributions of differences in ensemble spreads relative to RMSEs

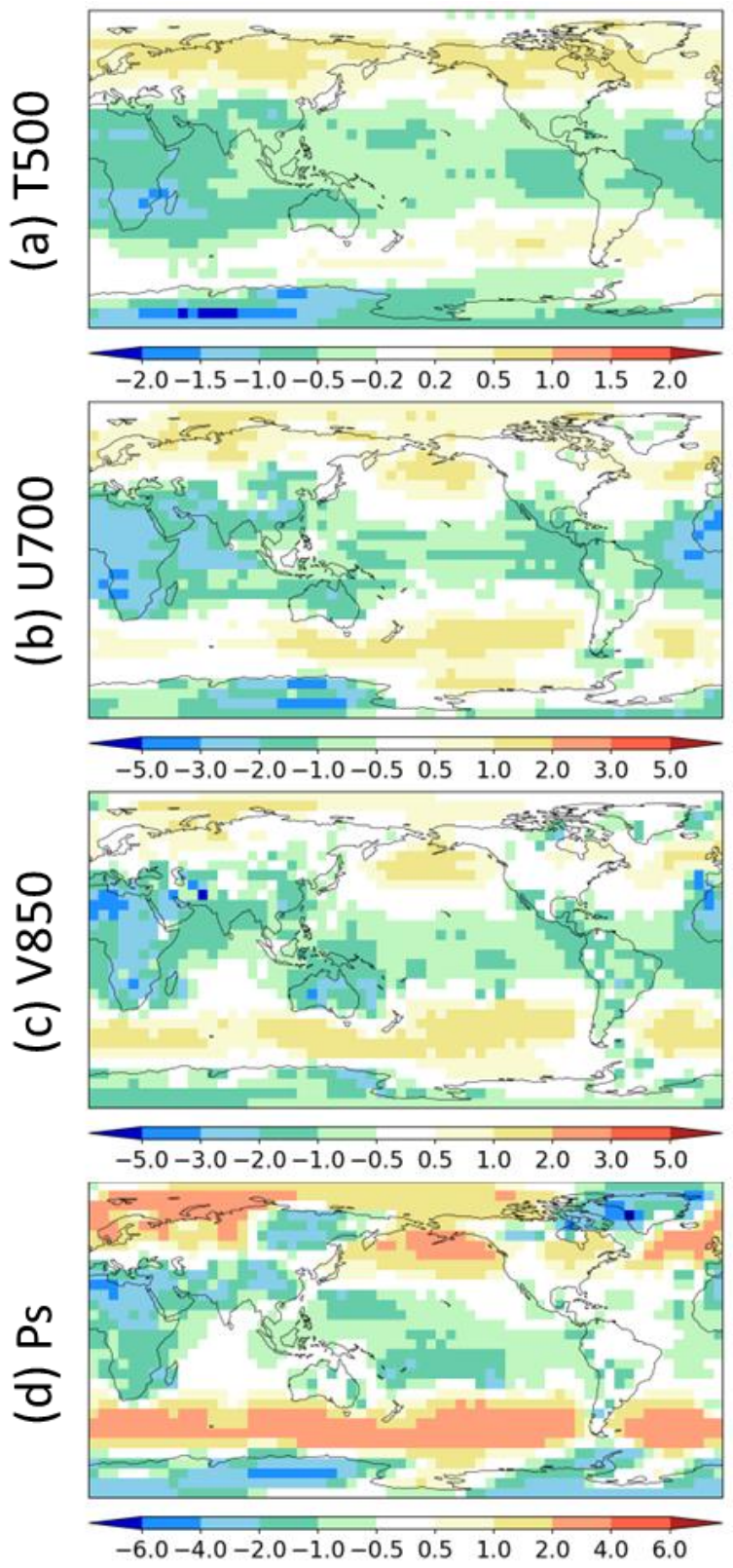

**Figure S6.** Mean spatial distributions of differences in ensemble spreads relative to RMSEs. Mean spreads and RMSEs are calculated over February 2016 – December 2017. Same variables for Figure 4, namely (a) temperature at 500 hPa, (b) zonal wind at 700 hPa, (c) meridional wind at 850 hPa, and (d) surface pressure.

## S7. Spatial distributions of differences in RMSEs and ensemble spreads of RTPS experiments from those of RTPP experiments

RMSEs and ensemble spreads of DA cycle experiments with RTPS ($\alpha = 1.20$) are compared to the results with RTPP ($\alpha = 0.90$). RTPS shows larger RMSEs in most areas and the differences are significant in the southern high-latitude area over oceans. In contrast, spreads are considerably smaller in those regions, while larger in the northern high-latitude area in U700 and V850.

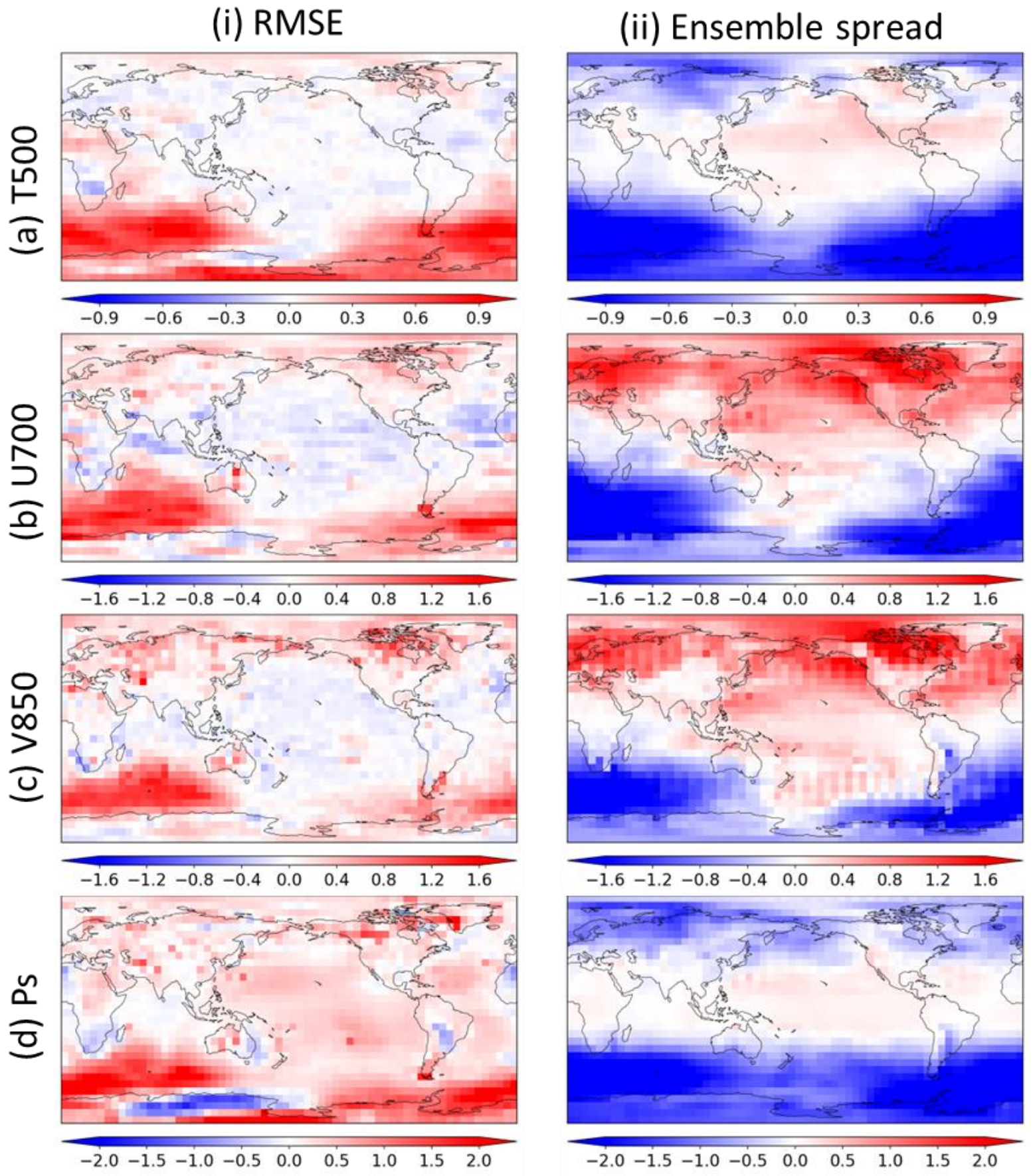

**Figure S7.** Differences of RMSEs and ensemble spreads of RTPS experiments from those of RTPP experiments. Same variables and span to Figure 4, i.e., (a) temperature at 500 hPa, (b) zonal wind at 700 hPa, (c) meridional wind at 850 hPa and (d) surface pressure over February 2016 – December 2017.

## S8. Data and software availability

Version 3.0.1 of the default ClimaX is preserved at https://doi.org/10.5281/zenodo.14258100 and developed openly at https://github.com/microsoft/ClimaX. The LETKF system is preserved at https://doi.org/10.5281/zenodo.14258014 and developed openly at https://github.com/skotsuki/speedy-lpf. WeatherBench data are available from https://mediatum.ub.tum.de/1524895 and NCEP observation data are from https://doi.org/10.5065/Z83F-N512. ClimaX-LETKF system and the scripts for reproducing this study is preserved at https://doi.org/10.5281/zenodo.15029841 and developed openly at

https://github.com/kotsuki-lab/Climax-LETKF. Main experiment data used to make the figures in this manuscript are preserved at https://doi.org/10.5281/zenodo.15016252 and WeatherBench data used for validation are preserved at https://doi.org/10.5281/zenodo.15029837. PREPBUFR format file decoding and observation thinning program and scripts for running them are preserved at https://doi.org/10.5281/zenodo.15029781 and developed openly at https://github.com/kotsuki-lab/PREPBUFR. Assimilated observation data are preserved at https://doi.org/10.5281/zenodo.15029720.